\documentclass[10pt,twocolumn]{article}

\usepackage[margin=1in]{geometry}
\usepackage{times}
\usepackage{graphicx}
\usepackage{amsmath}
\usepackage{amssymb}
\usepackage{booktabs}
\usepackage{multirow}
\usepackage{xcolor}
\usepackage{hyperref}
\usepackage{url}
\usepackage{cite}
\usepackage{caption}
\usepackage{subcaption}
\usepackage{enumitem}
\usepackage{microtype}
\usepackage{balance}
\usepackage[capitalise]{cleveref}

\hypersetup{
  colorlinks=true,
  linkcolor=blue!70!black,
  citecolor=green!50!black,
  urlcolor=blue!70!black
}

\title{\textbf{AIForge-Doc: A Benchmark for Detecting AI-Forged\\
Tampering in Financial and Form Documents}}

\author{
  Jiaqi Wu$^{*}$,
  Yuchen Zhou$^{*}$,
  Muduo Xu,
  Zisheng Liang,
  Simiao Ren$^{\dagger}$,\\
  Jiayu Xue,
  Meige Yang,
  Siying Chen,
  Jingheng Huan\\[0.5em]
  {\small $^{*}$Equal contribution \quad $^{\dagger}$Corresponding author}\\[0.3em]
  {\small Duke University: \texttt{\{jw933, yz946, zisheng.liang, siying.chen, jingheng.huan\}@duke.edu}}\\
  {\small New York University: \texttt{mx2336@nyu.edu} \quad
         University of North Carolina: \texttt{xuejiayu@unc.edu}}\\
  {\small Scam.ai: \texttt{benren@scam.ai} \quad University of Southern California: \texttt{maggieya@usc.edu}}
}

\date{}

\begin{document}
\maketitle
{\renewcommand{\thefootnote}{}\footnote{Authorship of this paper does not imply employment by or affiliation with Scam.AI (Reality Inc.); some authors are independent researchers who collaborated on this work.}\addtocounter{footnote}{-1}}

\begin{abstract}
We present \textbf{AIForge-Doc}, the first dedicated benchmark targeting
\emph{exclusively} diffusion-model-based inpainting in financial and form documents
with pixel-level annotation.
Existing document forgery datasets rely on traditional digital editing tools
(e.g., Adobe Photoshop, GIMP), creating a critical gap: state-of-the-art detectors
are \emph{blind} to the rapidly growing threat of AI-forged document fraud.
AIForge-Doc addresses this gap by systematically forging numeric fields in
real-world receipt and form images using two AI inpainting APIs---\textsc{Gemini 2.5
Flash Image} and \textsc{Ideogram v2 Edit}---yielding \textbf{4{,}061} forged images
from four public document datasets (CORD, WildReceipt, SROIE, XFUND) across nine
languages, annotated with pixel-precise tampered-region masks in DocTamper-compatible
format.
We benchmark three representative detectors---TruFor~\cite{guillaro2023trufor},
DocTamper~\cite{qu2023doctamper}, and a zero-shot GPT-4o judge---and find that all
existing methods degrade substantially:
TruFor achieves AUC=\textbf{0.751} (zero-shot, out-of-distribution)
vs.\ AUC=0.96 on NIST16;
DocTamper achieves AUC=\textbf{0.563} vs.\ AUC=0.98 in-distribution,
with pixel-level IoU=0.020;
GPT-4o achieves only \textbf{0.509}---essentially at chance---confirming that
AI-forged values are indistinguishable to automated detectors and VLMs.
These results demonstrate that AIForge-Doc represents a qualitatively new and unsolved
challenge for document forensics.

\end{abstract}

\section{Introduction}
\label{sec:intro}

\begin{figure*}[t]
  \centering
  \includegraphics[width=\textwidth]{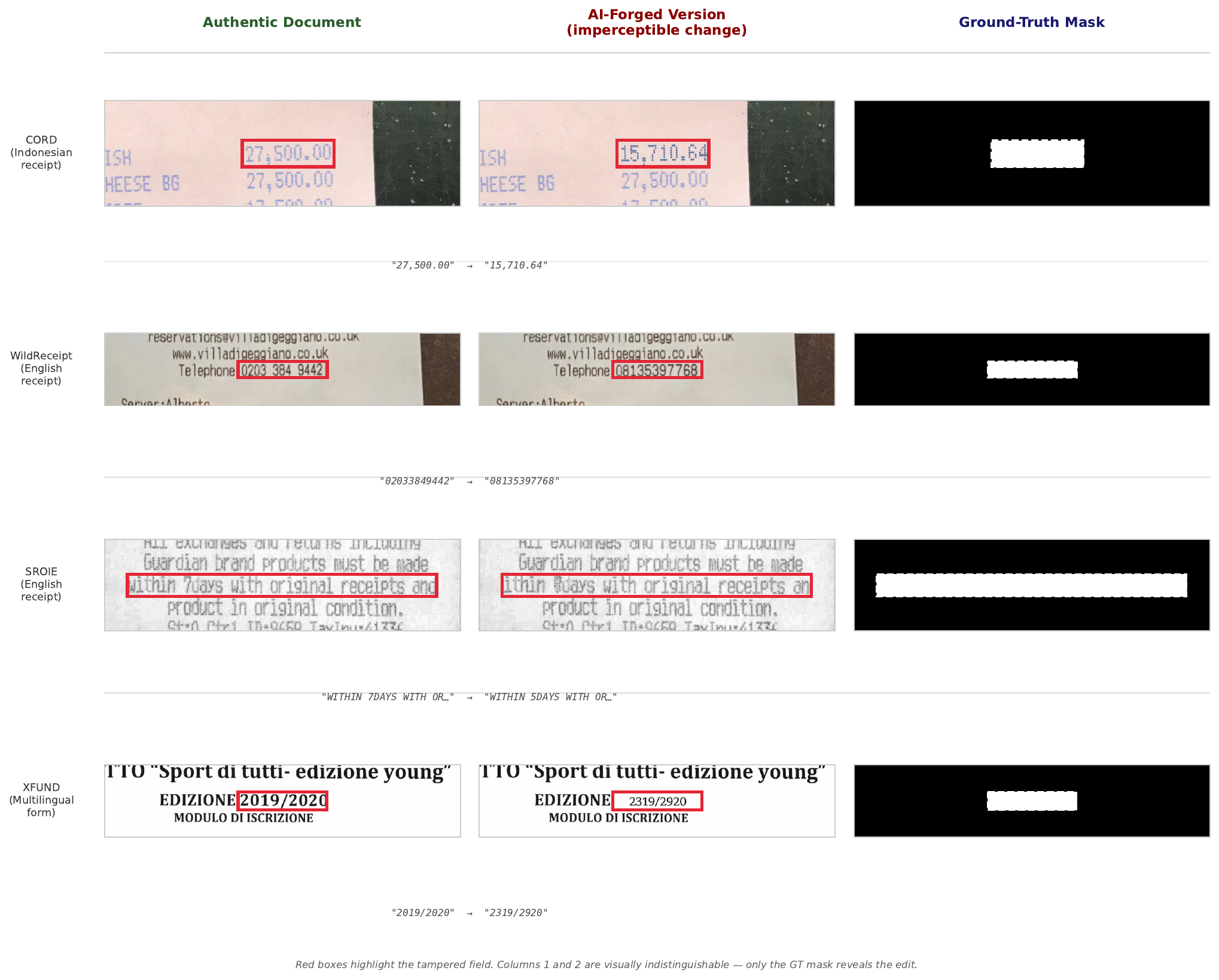}
  \caption{
    \textbf{AIForge-Doc: AI-inpainted document forgeries pass visual inspection and
    are difficult to distinguish from authentic documents.}
    Each row: authentic document with target field highlighted (left),
    AI-forged version (center), pixel-precise ground-truth mask (right).
    From top: CORD receipt (Ideogram v2 Edit), WildReceipt (Gemini 2.5 Flash Image),
    SROIE receipt (Gemini), XFUND multilingual form (Ideogram v2 Edit).
    The tampered region---a single numeric field---comprises a median of 0.9\% of
    image pixels, yet contains the forensically critical edit.
  }
  \label{fig:mask_examples}
\end{figure*}

Document fraud is an escalating global problem.
Digital document forgeries increased \textbf{244\%} year-over-year in 2024, and for the
first time digital forgeries (57\%) surpassed physical counterfeits as the dominant
fraud method~\cite{entrust2025fraud}---a 1,600\% increase since 2021.
A deepfake or AI-manipulated document attempt occurred every five minutes during 2024,
with generative AI tools cited as the primary enabler.
Historically, document forgery required expert knowledge of image editing software,
leaving characteristic traces---compression artifacts, cloning patterns, statistical
anomalies in noise residuals---that computational forensic methods can reliably detect.

The arrival of generative AI has fundamentally changed this threat landscape.
State-of-the-art diffusion-model-based inpainting APIs (e.g., Gemini 2.5 Flash Image,
Ideogram v2 Edit) can now convincingly replace a specific word or number in a
document photograph while seamlessly blending with the surrounding font, texture, and
background---all in under one second and at approximately \$0.01 per edit.
Unlike Photoshop-based edits, AI-forged regions exhibit no obvious compression seams or
cloning signatures; instead, the generator synthesizes plausible-looking pixels that are
statistically consistent with the original, making detection fundamentally harder.

\paragraph{The dataset gap.}
Despite the urgency of this threat, no public benchmark exists for evaluating detectors
against AI-forged document tampering.
The leading document-specific forgery datasets---DocTamper~\cite{qu2023doctamper} (170k images),
RTM~\cite{luo2024rtm} (9k images), and the ICDAR~2023 TII benchmark~\cite{icdar2023dtti}
(11k images)---all use traditional copy-move, splicing, or typesetting manipulation.
Even the most recent OSTF benchmark~\cite{qu2025ostf}, while including diffusion-based
methods, focuses on scene-text images (storefronts, signs) rather than financial or form
documents, and uses bounding-box rather than pixel-level mask annotation.
Crucially, even if a detector were trained on OSTF-style scene-text AI forgeries, there
is no guarantee it would generalize to the different visual domain and targeted numeric
manipulation of financial receipts and form documents---a gap our empirical evaluation
(§\ref{sec:results}) confirms.
General-purpose image forgery datasets (COVERAGE~\cite{wen2016coverage},
Columbia~\cite{columbia2004}, CASIA~\cite{dong2013casia}, NIST16~\cite{guan2019mfc}) are
similarly free of AI-generated content.
Detectors trained on these corpora are therefore evaluated exclusively on the distribution
they were designed for, with no guarantee of robustness to the qualitatively different
artifacts produced by neural inpainting in financial documents.

\paragraph{Our contribution.}
We introduce \textbf{AIForge-Doc}, the first benchmark \emph{targeting exclusively}
diffusion-model-based inpainting in financial and form documents with pixel-level annotation,
with three core contributions:

\begin{enumerate}[leftmargin=*, topsep=2pt, itemsep=1pt]
  \item \textbf{A novel dataset of AI-forged document images.} We systematically tamper
    with numeric fields in 4{,}061 source images from four public datasets using two AI
    inpainting APIs (Gemini 2.5 Flash Image and Ideogram v2 Edit), selected from seven
    evaluated systems (five rejected for garbled text output or insufficient resolution),
    yielding 4{,}061 forged images with pixel-precise ground-truth masks in an 80/20
    train/test split.
    Field types vary by dataset: financial amounts (CORD), telephone and address numerics
    (WildReceipt), text-embedded numbers (SROIE), and form answer fields (XFUND).
    The training partition (3{,}249 images) is intended as a resource for the community
    to develop and train future detectors specifically targeting AI-inpainting artifacts.

  \item \textbf{A reproducible generation pipeline.} We release a fully automated
    open-source pipeline that ingests raw source documents, selects high-priority numeric
    fields, generates plausible alternative values, runs context-window inpainting via
    local or API-accessed models, and packages results in DocTamper-compatible format.

  \item \textbf{A baseline evaluation revealing a critical detection gap.} We benchmark
    TruFor~\cite{guillaro2023trufor}, DocTamper~\cite{qu2023doctamper}, and GPT-4o zero-shot on
    AIForge-Doc and find substantial degradation:
    TruFor achieves AUC=0.751 on AIForge-Doc (zero-shot) vs.\ 0.96 on NIST16 (per original
    authors); DocTamper achieves AUC=0.563 on AIForge-Doc vs.\ 0.98 on its own
    in-distribution test set, with IoU falling from 0.71 to 0.020; GPT-4o achieves only
    0.509---essentially random.
    This establishes AI-forged document tampering as an open and important research problem.
\end{enumerate}

\paragraph{Scope and limitations.}
AIForge-Doc focuses on \emph{localized numeric field forgery} in receipts and form
documents---the highest-risk scenario for financial fraud.
We do not cover wholesale document synthesis (e.g., GAN-generated identity documents
from scratch), signature forgery, or LLM-generated textual fraud, which we consider
complementary and leave for future work.

\section{Related Work}
\label{sec:related}

\subsection{Document Forgery Datasets}

\paragraph{Document text tampering datasets.}
DocTamper~\cite{qu2023doctamper} (CVPR 2023) is the largest prior work: 170,000 document
images (contracts, invoices, receipts) with character- and word-level text substitution,
insertion, and deletion---all traditional typesetting edits with no AI-generated content.
RTM~\cite{luo2024rtm} (\emph{Pattern Recognition} 2024) provides 9,000 images including
6,000 manually tampered by professional editors, revealing that detectors trained on
synthetic forgeries fail on human-crafted real-world ones.
The ICDAR 2023 TII dataset~\cite{icdar2023dtti} adds 11,385 images across classification
and localization tracks.

\paragraph{OSTF and AI-inpainting benchmarks.}
OSTF~\cite{qu2025ostf} (AAAI 2025) is the closest predecessor, assembling 4,418 images
by replacing text regions with outputs from eight methods (four diffusion-based, four
conventional) in natural scene-text photographs (storefronts, signs, menus).
AIForge-Doc differs from OSTF along four axes.
\emph{Document type}: OSTF targets open-domain scene-text; AIForge-Doc exclusively
targets financial receipts and structured forms, where numeric field tampering carries
direct fraud consequences---a forged total on a receipt constitutes financial fraud
in a way that a wrong number on a storefront sign does not.
\emph{Threat model}: OSTF benchmarks a broad spectrum of manipulation methods (4 diffusion
+ 4 conventional); AIForge-Doc isolates the specific threat posed by consumer-accessible
diffusion APIs (Gemini 2.5 Flash Image, Ideogram v2 Edit), where a non-expert can forge
a document in under one second for \$0.01.
\emph{Annotation format}: OSTF uses word-level polygon masks and reports bounding-box
AP metrics; AIForge-Doc provides full-image binary masks in DocTamper-compatible format,
enabling direct IoU/AUC-based comparison with existing document forensics methods.
\emph{Generalization gap}: Most critically, OSTF demonstrates that AI text replacement
\emph{is} detectable when detectors are trained on the appropriate manipulation
distribution.
AIForge-Doc asks a harder and distinct question: \emph{does knowledge of AI scene-text
forgeries transfer to targeted financial document forgery?}
Our zero-shot evaluation (\cref{sec:results}) provides evidence that it does not---
TruFor achieves AUC\,=\,0.751 and DocTamper AUC\,=\,0.563 despite strong performance
on their respective training distributions---establishing financial document forgery as
an open research problem even given the existence of OSTF-style training data.
AIForge-Doc is also publicly released without registration.
FGDTD~\cite{fgdtd2025} (2025) introduces fine-grained tampering classification over
16,479 images across 12 source datasets.
SAGI~\cite{sagi2025} provides 95,000+ AI-inpainted images across multiple diffusion
pipelines; AIForge-Doc complements it by focusing specifically on document content
and targeted numeric-field manipulation.

\paragraph{General image forgery datasets.}
Columbia~\cite{columbia2004}, COVERAGE~\cite{wen2016coverage}, CASIA~\cite{dong2013casia},
NIST MFC~\cite{guan2019mfc}, and FF++~\cite{rossler2019faceforensics} are standard
benchmarks for copy-move, splicing, and deepfake detection in natural photographs;
none target document images or AI-based inpainting.

\subsection{Tampering Detection Methods}

\paragraph{General forensic detectors.}
ManTraNet~\cite{wu2019mantranet}, CAT-Net~\cite{kwon2021catnet}, PSCC-Net~\cite{liu2022psccnet},
HiFi-Net~\cite{guo2023hifinet}, and IML-ViT~\cite{ma2023imlvit} represent the range of
general-purpose image forgery detectors.
Ren et al.~\cite{ren2025deepfake} demonstrate that state-of-the-art deepfake detectors
frequently fail to generalize to real-world conditions outside their training distribution,
a finding directly echoed in our zero-shot evaluation on AI-forged document content.
TruFor~\cite{guillaro2023trufor} (CVPR 2023) is the current state of the art, combining
a transformer backbone with NoisePrint++, a learnable camera-model fingerprint, to
achieve strong generalization across manipulation types; we use it as our primary baseline.

\paragraph{Document-specific detectors.}
DocTamper's detector~\cite{qu2023doctamper} is the only published method trained
specifically on document forgeries.
It incorporates document-specific priors via frequency-domain auxiliary losses,
achieving substantial gains over general detectors on its own test set.
We evaluate it zero-shot on AIForge-Doc to measure the generalization gap to AI-forged content.

\paragraph{LLM/VLM as forensic judges.}
\cite{ren2025llmdeepfake} benchmark multimodal LLMs (including GPT-4o and Gemini)
against traditional deepfake detectors.
\cite{liang2025llmdocmanip} extend this to fraudulent document detection via prompt
optimization---the closest precedent to our GPT-4o baseline.
We include GPT-4o zero-shot to measure what world-knowledge reasoning can achieve without
labeled training data.

\subsection{AI Inpainting and Its Forensic Signatures}

Diffusion-model-based inpainting~\cite{rombach2022ldm,flux2024} achieves photorealistic
local edits by conditioning denoising on a masked region.
\cite{corvi2023detection} and \cite{ricker2024aeroblade} study detection of
diffusion-generated images at the full-image level.
Ren et al.~\cite{ren2026aigendet} provide a comprehensive zero-shot benchmark of
16 open-source AI-generated image detectors across 12 datasets, finding no universal
winner---a breadth finding complementary to our targeted in-document localization study.
Detection of \emph{localized} diffusion inpainting within otherwise authentic documents
is far less studied and represents the core challenge in AIForge-Doc.

\section{Dataset Construction}
\label{sec:construction}

\begin{figure*}[t]
  \centering
  \begin{subfigure}{\textwidth}
    \centering
    \includegraphics[height=7.5cm,keepaspectratio]{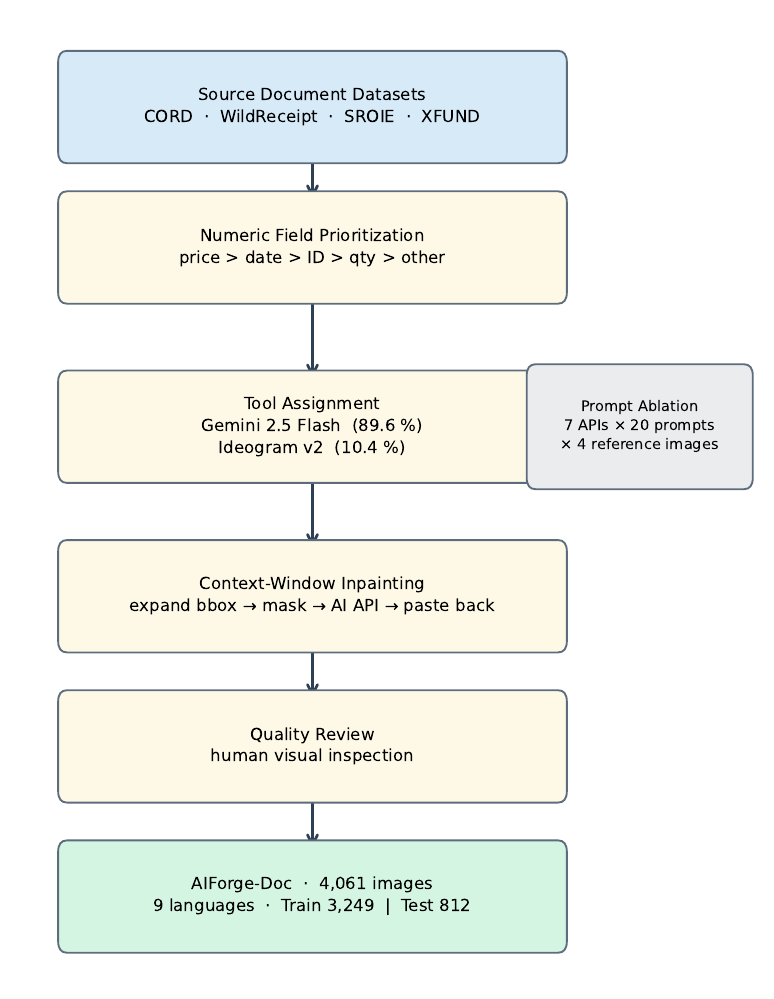}
    \caption{Overall dataset creation workflow.}
    \label{fig:flowchart}
  \end{subfigure}

  \vspace{6pt}

  \begin{subfigure}{\textwidth}
    \centering
    \includegraphics[width=\textwidth]{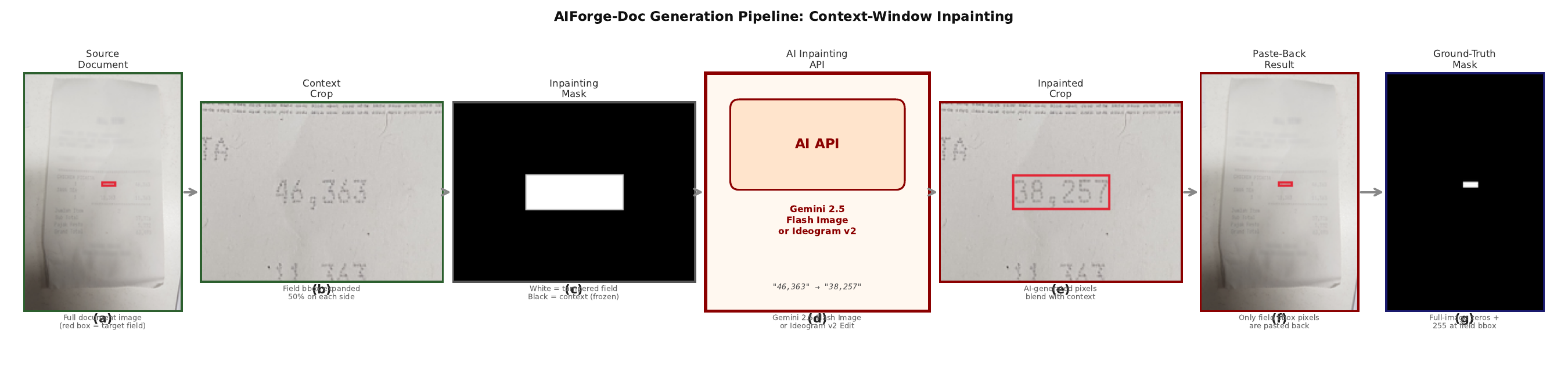}
    \caption{Per-image context-window inpainting technique (steps a--f).}
    \label{fig:pipeline_detail}
  \end{subfigure}
  \caption{
    \textbf{AIForge-Doc generation pipeline.}
    \emph{Top}: the overall dataset creation workflow, from source datasets through
    field selection and tool assignment (informed by a 320-trial prompt ablation study)
    to the final DocTamper-compatible dataset.
    \emph{Bottom}: the per-image context-window inpainting technique---starting from a
    source document (a), we expand a context crop (b), create a binary inpainting mask
    (c), feed the crop and mask to the AI API (d), and paste only the field region back
    into the full image (e) to produce the forged output and pixel-precise ground-truth
    mask (f).
  }
  \label{fig:pipeline}
\end{figure*}

\subsection{Source Document Datasets}
\label{sec:sources}

AIForge-Doc is built on top of four publicly available document datasets spanning receipts
and forms in multiple languages (\cref{tab:sources}).

\paragraph{CORD v2~\cite{park2019cord}.}
The Consolidated Receipt Dataset contains 1,000 scanned receipt images in Indonesian
with comprehensive key-value annotations for ${\sim}$30 field types including
item names, unit prices, subtotals, and total prices.
We use the full 1,000-image split, prioritizing \texttt{menu.price},
\texttt{total.total\_price}, and \texttt{sub\_total.subtotal\_price} fields.

\paragraph{WildReceipt~\cite{sun2021wildreceipt}.}
An open-domain English receipt dataset with approximately 1,740 images collected from
diverse scanners and mobile cameras, annotated with 25 key-value field types.
We use 1,696 images from the released split.
Financial amount fields are labeled with IDs 13--14 (product price), 19--20 (tax), and
23--24 (total).

\paragraph{SROIE~\cite{huang2019sroie}.}
The ICDAR 2019 Scanned Receipt OCR and Information Extraction dataset contains 626 training
and 347 test receipt images with four key fields: \textit{company}, \textit{date},
\textit{address}, and \textit{total}.
We target numeric fields, prioritizing \textit{total} (financial amount) and
\textit{date}; when these are unavailable the field selector falls back to
address-embedded numerics and policy-text fields containing numeric values
(e.g., return windows such as ``WITHIN 7 DAYS''), using 946 images.

\paragraph{XFUND~\cite{xu2022xfund}.}
A multilingual form understanding dataset in seven non-English languages (ZH, JA, ES, FR,
IT, DE, PT) with key-value annotations.
We target numeric \texttt{answer} entities, contributing 419 images.
XFUND forms are high-resolution A4 scans (up to 4961$\times$7016 px at 600 dpi).

\subsection{Field Selection and Value Mutation}
\label{sec:mutation}

\paragraph{Field prioritization.}
For each source image, we select the single highest-priority numeric field according to:
(1) financial amount, (2) date, (3) document ID/number, (4) quantity, (5) other numeric.
This focuses the dataset on the highest-risk forgery scenarios for financial fraud.

\paragraph{Value mutation.}
We generate plausible alternative values that look realistic but differ from the original:

\begin{itemize}[leftmargin=*, topsep=2pt, itemsep=1pt]
  \item \textbf{Monetary fields:} Multiply by a random scale factor drawn from
    $\mathcal{U}(1.15, 3.0)$ (scale up) or $\mathcal{U}(0.20, 0.85)$ (scale down),
    applied with 50\% probability each, producing a roughly symmetric distribution of
    scaled-up and scaled-down forgeries.
    The mutation loop retries up to 30 times to find a value with the same character
    count as the original, minimizing visual discrepancy in fixed-width fonts; if
    unsuccessful, digit-flipping is used as a fallback (which preserves length by
    construction).

  \item \textbf{Date fields:} Perturb year by $\pm$1--5, month by $\pm$1--3, or
    day by $\pm$3--15, with calendar validity enforced.

  \item \textbf{Document IDs:} Flip 1--2 randomly selected digits.

  \item \textbf{Quantities:} Multiply by 2--5$\times$.
\end{itemize}

\subsection{Context-Window Inpainting Technique}
\label{sec:inpaint_technique}

A critical design choice is the \emph{context-window} approach illustrated in
\cref{fig:pipeline}.
Naïvely feeding an entire document image to an inpainting model (a) risks global drift:
the model may hallucinate backgrounds or alter surrounding text.
Instead, we:

\begin{enumerate}[leftmargin=*, topsep=2pt, itemsep=1pt]
  \item Expand the field bounding box by 50\% on each side (minimum 150 px padding) to
    create a \emph{context crop} that includes neighboring characters for font reference.
  \item Create a binary mask on the context crop: white (255) over the field, black (0)
    elsewhere.
  \item Run inpainting on the context crop only with a font-preserving prompt
    (\cref{sec:prompt}).
  \item Paste \emph{only the field region} (original bbox, no padding) from the inpainted
    context back into the full-resolution source image.
  \item Generate the ground-truth mask as a full-image binary PNG with 255 at the exact
    field bbox and 0 everywhere else.
\end{enumerate}

This guarantees that (i) the model sees surrounding text for font/color reference,
(ii) only pixels within the field bbox are altered in the final image, and (iii) the
ground-truth mask is pixel-perfect.
The paste-back step (step~4) is the key safeguard: regardless of what the inpainting
model does to surrounding pixels in the context crop, only the exact field bbox region
is ever written back to the source image, ensuring mask accuracy by construction rather
than by relying on the model's compliance with the prompt.
To verify that Gemini's editing does not affect regions outside the declared bbox,
we compared non-bbox pixels between originals and forged images across 200 randomly
sampled Gemini outputs and found that the context-window paste-back step eliminates
any out-of-bbox drift by construction---since only the exact field region is pasted
from the inpainted context into the full image.

\paragraph{Inpainting prompt.}
\label{sec:prompt}
All tools receive the following templated prompt:

\begin{quote}\small\itshape
``Change only the text in the masked region to read `\{new\_value\}'.
Preserve the exact font family, weight, size, letter-spacing, color, and vertical
baseline of the surrounding characters on the same line.
Do not alter the background texture, borders, lines, or any other element.
The edit must be visually indistinguishable from the original document.''
\end{quote}

\subsection{AI Inpainting Tools}
\label{sec:tools}

We prototyped seven AI inpainting systems and deployed two for mass generation.

\paragraph{Deployed tools (2).}
\begin{itemize}[leftmargin=*, topsep=2pt, itemsep=1pt]
  \item \textbf{Gemini 2.5 Flash Image} (OpenRouter): Google's multimodal model with
    image generation capability, accessed via \texttt{google/gemini-2.5-flash-image}.
    Gemini does not accept a binary mask tensor; instead we draw a 3-pixel bright-green
    rectangle on the context crop to mark the edit region and reference it in the prompt.
  \item \textbf{Ideogram v2 Edit} (fal.ai): A high-fidelity text-rendering
    inpainting system~\cite{ideogram2024}, particularly suited for typographic content.
    Accessed via \texttt{fal\mbox{-}ai/\allowbreak ideogram/\allowbreak v2/\allowbreak edit}.
\end{itemize}

\paragraph{Evaluated but disabled tools (5).}
We systematically evaluated five additional inpainting systems through a prompt
ablation study (\cref{sec:ablation}; \cref{fig:prompt_ablation}): for each
API-accessible tool, we tested 20 diverse prompt formulations on four reference
images across all source datasets (320 trials total).

\textbf{FLUX Fill Pro}~\cite{flux2024}: generates plausible digit shapes but
consistently \emph{wrong values} (e.g., ``112,800'' or ``155,700'' instead of
target ``196,718''), with inconsistent font weight across all 20 prompt variants.

\textbf{SD 3.5 Medium}~\cite{sd35}: the worst performer; frequently renders
the \emph{prompt text itself} into the image (e.g., ``Replace Text,''
``Restore Beautifully'') or produces garbled characters and emoji, ignoring the
inpainting task entirely regardless of prompt formulation.

\textbf{GPT-Image-1}~\cite{gptimage1}: falls back to DALL-E-2~\cite{ramesh2022dalle2}
at 512$\times$512~px; occasionally produces correct numerals but with mismatched
font size, excessive boldness, and blurry upscaling artifacts that fail human review.

\textbf{SD 1.5 Inpainting}~\cite{rombach2022ldm}: 512$\times$512 native resolution
produces uniform blurry patches; text is entirely unreadable at document scale.

The remaining tool, \textbf{FLUX Fill Dev}~\cite{flux2024}, requires 24~GB VRAM
and was not available for mass generation; it shares the FLUX architecture with
Fill Pro and similarly lacks multimodal text understanding.

Across all 320 ablation trials, \emph{zero} outputs from any rejected tool met
our quality bar.
The failure is architectural: only models with multimodal language understanding
(Gemini 2.5 Flash Image, Ideogram v2 Edit) can perform character-accurate text
replacement in document images.
AIForge-Doc therefore concentrates on these two tools.
Pipeline code and ablation results for all seven tools are released for
reproducibility.

\subsection{Engineering Adaptations}
\label{sec:engineering}

Several non-trivial implementation challenges arose during large-scale generation;
we document them here for reproducibility.

\paragraph{Gemini mask encoding.}
As noted above, Gemini does not accept a binary mask tensor.
To improve text legibility on small fields, we 2$\times$ upsample the context crop
(LANCZOS) before sending to the API and downsample the result before pasting back.

\paragraph{Ideogram content filtering.}
The Ideogram v2 Edit API applies an internal content safety checker that flagged
83 requests ($\approx$16.6\% of Ideogram specs), predominantly from SROIE receipts
containing real Malaysian business addresses and registration numbers, and XFUND
forms with government document content.
Our pipeline distinguishes \emph{content-filter failures} (matched on
\texttt{content\_policy\_violation}) from ordinary API errors: content-filtered
requests are immediately skipped and logged to \texttt{skipped\_specs.jsonl};
ordinary errors trigger exponential-backoff retry (3 attempts: 5 s, 10 s, 20 s);
budget-exhaustion errors halt generation entirely.
All 81 unique skipped Ideogram specs were resubmitted to Gemini and completed
successfully, yielding zero net data loss.
The 81 rerouted specs are labeled in metadata with \texttt{assigned\_tool=gemini}
and are excluded from the Ideogram subset in per-tool analysis.

\paragraph{Separator hint in prompt.}
Early Ideogram outputs frequently omitted thousand-separator commas
(e.g., ``75000'' instead of ``75,000'').
We append a separator hint to the prompt: ``Use \{sep\} as the thousands separator,''
where \{sep\} is inferred from the original value.

\paragraph{Value length matching.}
When the mutated value has a different character count from the original, the inpainting
model must render more or fewer glyphs into the same bounding box, which degrades quality.
As described in \cref{sec:mutation}, the mutation loop retries up to 30 times to find a
length-matched value before falling back to digit-flipping.

\subsection{Quality Control}
\label{sec:quality}

Quality control proceeds in three stages applied to all 4,061 generated images:

\begin{enumerate}[leftmargin=*, topsep=2pt, itemsep=1pt]
  \item \textbf{Automated OCR check (all images).}
    As a coarse pre-filter, each forged image is validated with
    PaddleOCR~\cite{paddleocr}: we crop the tampered bbox region and verify that the
    OCR output contains at least one recognized token.
    Images where the model produced a blank, smeared, or hallucinated result are flagged
    and excluded; this step removes obviously broken outputs before human review.

  \item \textbf{Human review (all images).}
    Every generated image was reviewed by the authors via side-by-side preview panels
    (original vs.\ forged, with the tampered bbox highlighted)---the primary quality gate.
    Images with blank, garbled, or visually incoherent output are flagged immediately;
    all 4,061 final images passed this review.

  \item \textbf{Semantic plausibility check (metadata validation).}
    We verify that the forged value differs from the original value for every image,
    filtering any edge cases where the mutation loop produced an identical value.
\end{enumerate}

This three-stage pipeline complements the upstream tool screening (\cref{sec:tools}),
where 5 of 7 prototyped models were eliminated because they consistently produced
garbled or illegible text output---ensuring that AIForge-Doc contains only forgeries
at the quality frontier of current AI inpainting capability.
A final manual pass over the complete 812-image testing set confirmed legible,
realistic-appearing text in every case.

\subsection{Prompt Ablation Study}
\label{sec:ablation}

A natural concern is whether the five rejected tools (\cref{sec:tools}) might
succeed under different prompt formulations.
To address this, we conducted a systematic prompt ablation: for each of the four
API-accessible rejected tools (FLUX Fill Pro, GPT-Image-1, SD~3.5 Medium, and
SD~1.5 Inpainting), we tested \textbf{20 diverse prompts} on \textbf{four reference
images}---one from each source dataset (CORD, SROIE, XFUND, WildReceipt)---where
Gemini 2.5 Flash Image had already produced high-quality forgeries, yielding
$4 \times 20 \times 4 = 320$ inpainting trials total.

\paragraph{Prompt diversity.}
The 20 prompts span a wide spectrum of strategies:
(i)~\emph{minimal} (bare target value only),
(ii)~\emph{imperative} (``Replace masked text with \ldots''),
(iii)~our \emph{production template} used for mass generation,
(iv)~\emph{character-by-character} spelling of the target value,
(v)~\emph{chain-of-thought} (analyze surrounding typography, then act),
(vi)~\emph{role-play} variants (document restoration specialist,
     Photoshop retoucher, forensic examiner),
(vii)~\emph{typography expert} (match typeface, kerning, leading, stroke weight),
(viii)~\emph{negative constraints} (``Do NOT change pixels outside the mask''),
and (ix)~\emph{verbose exhaustive} (8-point numbered requirements list).
The full prompt set is provided in our released code.

\begin{figure*}[t]
  \centering
  \includegraphics[width=\textwidth]{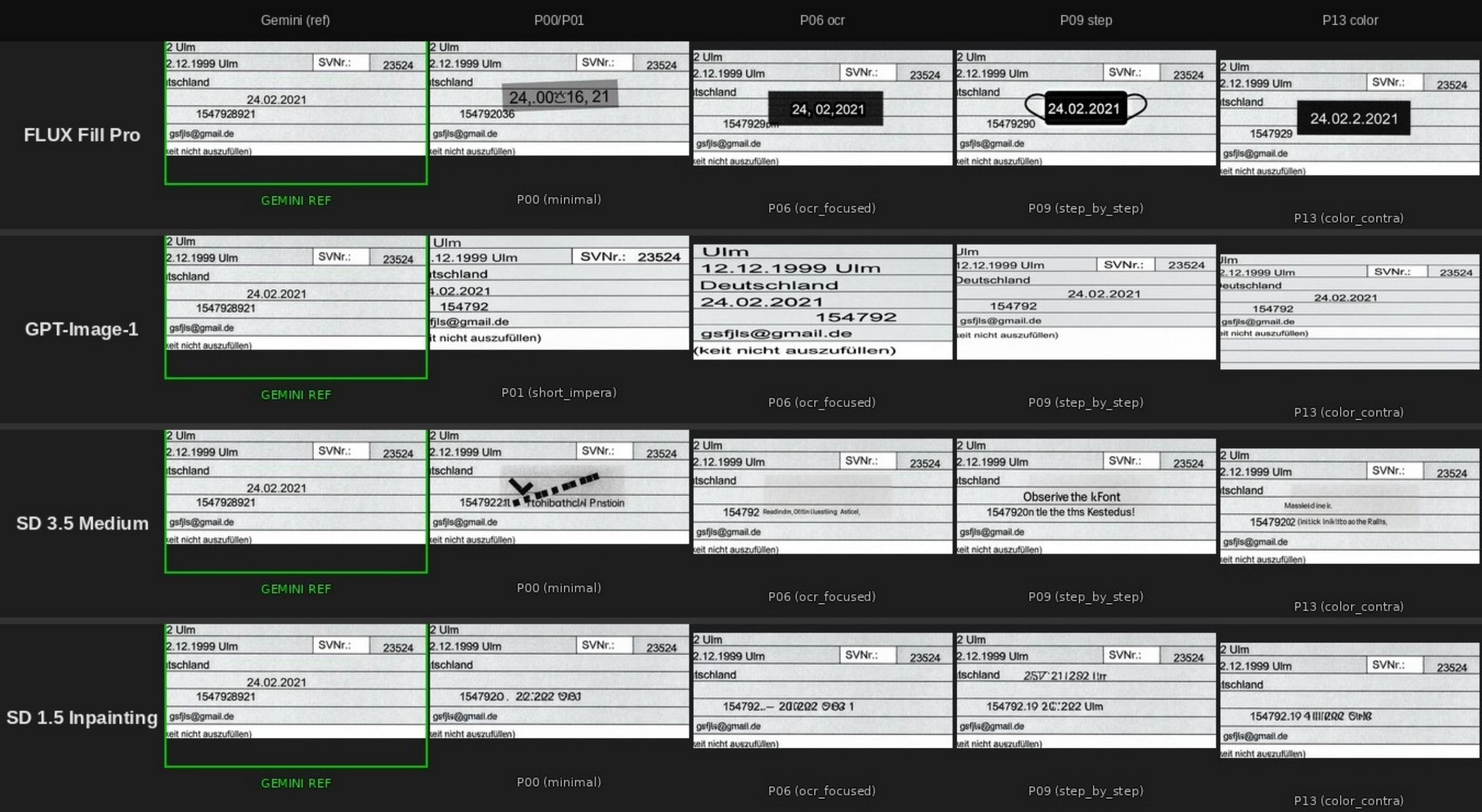}
  \caption{
    \textbf{Prompt ablation: 5 representative outputs per rejected tool on WildReceipt
    image 000002699 (\texttt{Prod\_item\_key}, ``HULAHAWAIIANT1''$\to$``HULAHAWAIIANT4'').}
    Each row shows the Gemini~2.5 Flash reference (green border) alongside outputs from
    four prompt strategies spanning minimal, OCR-focused, step-by-step, and color-aware
    prompts.
    FLUX Fill Pro renders plausible digit shapes but consistently wrong values;
    GPT-Image-1 produces a black N/A patch or blurry wrong-font text at 512~px;
    SD~3.5 Medium renders prompt text literally or produces garbled symbols;
    SD~1.5 Inpainting yields illegible blurred patches.
    No prompt strategy succeeds for any rejected tool.
    Full 20-prompt-variant grids are in \cref{fig:ablation_flux,fig:ablation_gpt,fig:ablation_sd35,fig:ablation_sd15} (Appendix~\ref{app:ablation}).
  }
  \label{fig:prompt_ablation}
\end{figure*}

\paragraph{Results.}
\Cref{fig:prompt_ablation} shows representative outputs.
None of the 320 trials produced output meeting our quality bar.
Crucially, \emph{no prompt strategy}---including detailed chain-of-thought
reasoning, domain-expert role-play, and verbose multi-constraint
specifications---overcame these limitations.
The two deployed tools succeed because they possess multimodal language
understanding that enables character-accurate text rendering; the rejected tools,
as pure image-synthesis diffusion models, lack this capability regardless of
prompt engineering.

\section{Dataset Statistics and Analysis}
\label{sec:statistics}

\begin{table}[t]
  \centering
  \caption{Source datasets in AIForge-Doc.}
  \label{tab:sources}
  \footnotesize
  \setlength{\tabcolsep}{3.5pt}
  \begin{tabular}{@{}lcccc@{}}
    \toprule
    Dataset & Doc Type & Lang & Images & Fields \\
    \midrule
    CORD v2~\cite{park2019cord}           & Receipt & id    & 1{,}000 & price, total \\
    WildReceipt~\cite{sun2021wildreceipt} & Receipt & en    & 1{,}696 & tel, addr, price \\
    SROIE~\cite{huang2019sroie}           & Receipt & en    & 946     & total, date \\
    XFUND~\cite{xu2022xfund}              & Form    & multi & 419     & answer (num) \\
    \midrule
    \textbf{Total}                         & ---     & ---   & \textbf{4{,}061} & --- \\
    \bottomrule
  \end{tabular}
\end{table}

\subsection{Scale and Composition}

AIForge-Doc contains \textbf{4{,}061} forged images derived from 4,061 source images across
four receipt and form datasets (\cref{tab:sources}).
Each source image contributes exactly one forged image (single field, single tool),
maintaining a clean one-to-one correspondence between authentic and forged examples.

\paragraph{Field type distribution.}
Field selection follows the prioritization scheme in \cref{sec:mutation}
(financial amount $>$ date $>$ document ID $>$ quantity $>$ other numeric).
In practice, the selected field type is determined by each source dataset's annotation
structure, and the distribution reflects their coverage rather than the priority ordering:
CORD images predominantly contribute financial amount fields (\texttt{menu.price},
\texttt{total.total\_price}, \texttt{sub\_total.subtotal\_price}), which are richly
annotated.
WildReceipt has limited financial amount annotations accessible at the field level;
the priority scheme therefore falls back to other labeled numeric fields---primarily
telephone numbers (\texttt{Telephone\_key}, $n{=}931$) and store address numerics
(\texttt{Store\_addr\_key}, $n{=}457$)---which are the most abundant numeric annotations
in that dataset.
SROIE's available annotations are contact and return-policy text fields containing
embedded numeric values (e.g., ``WITHIN 7 DAYS WITH RECEIPTS''), along with a subset
of date fields.
XFUND images contribute numeric \texttt{answer} fields from multilingual forms.
While the dataset therefore spans a broader range of document numeric fields than
pure financial amounts, every selected field is a plausible target for document
fraud: forging a business telephone number, address code, or policy date can
mislead document verification just as a forged total can.
The dataset is thus best characterized as an \emph{AI-forged document field} benchmark,
with CORD providing the most forensically critical financial amount examples.

\paragraph{Tool distribution.}
Gemini 2.5 Flash Image accounts for 89.6\% of forgeries (3{,}639 images)
and Ideogram v2 Edit for 10.4\% (422 images).
The primary driver of this imbalance is \emph{output quality}: we prioritize
dataset quality over tool balance, and Ideogram v2 Edit was restricted to the
CORD and WildReceipt subsets where its typographic output consistently met our
quality bar.
On SROIE and XFUND content, Ideogram's content safety filter rejected
83 requests ($\approx$16.6\% of its allocated specs, see \cref{sec:engineering}),
and the remaining outputs on those document types showed lower typographic
fidelity than Gemini; those specs were therefore rerouted to Gemini.
Ideogram's per-image cost (2$\times$ Gemini's) further reinforced the decision
to concentrate Gemini usage where Ideogram underperformed.
The 20/80 stratified split yields 85 Ideogram and 727 Gemini images in the test set,
reflecting this quality-driven composition.
This imbalance has forensic significance: the two tools leave detectably different
artifact signatures, and the asymmetric test-set ratio must be accounted for when
interpreting per-tool detection results (\cref{sec:results}).

\paragraph{Data partitions.}
We release a fixed 80/20 partition (3{,}249 / 812 images), stratified by
source dataset and tool.
All baseline experiments in \cref{sec:results} are zero-shot evaluations on the
test partition.

\subsection{Mask Format}

Ground-truth masks follow the DocTamper convention:
\begin{itemize}[leftmargin=*, topsep=2pt, itemsep=1pt]
  \item 8-bit grayscale PNG, same resolution as the source image.
  \item Pixel value 0: authentic region.
  \item Pixel value 255: tampered region.
  \item Tampered region = the exact field bounding box (tight, no padding).
\end{itemize}

\Cref{fig:mask_examples} (page 1) shows representative authentic/forged image pairs with
their ground-truth masks across four source datasets.

\subsection{Tampered Region Analysis}
\label{sec:region_stats}

\begin{figure}[t]
  \centering
  \includegraphics[width=\linewidth]{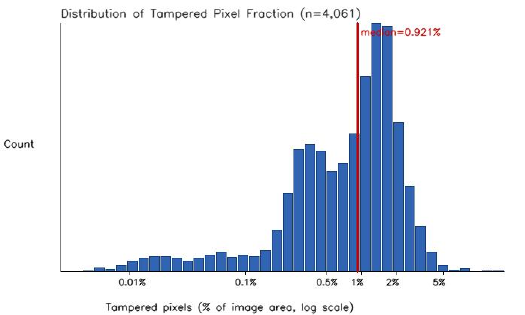}
  \caption{
    \textbf{Distribution of tampered pixel fraction across 4,061 AIForge-Doc images.}
    Median: 0.92\% (vertical blue line); IQR: [0.35\%, 1.55\%].
    Over 99\% of pixels in each image are unmodified---the tampered region
    is a small, localized field bbox.
  }
  \label{fig:region_dist}
\end{figure}

The tampered region in each AIForge-Doc image is a single numeric field bounding
box---a highly localized edit within an otherwise authentic document.
\Cref{fig:region_dist} shows the distribution of tampered pixel fraction
(bbox area $\div$ total image area) across all 4{,}061 images.
The median tampered area is \textbf{5{,}589~px$^2$}, comprising a median of
\textbf{0.92\%} of total image pixels (IQR: [0.35\%, 1.55\%]);
52\% of images have less than 1\% of pixels tampered.
XFUND forms have the smallest relative tamper fraction due to their high resolution
(up to 4{,}961$\times$7{,}016 px at 600 dpi).
This extreme spatial sparsity---over 99\% of pixels are unmodified---means
detection is analogous to finding a needle in a haystack, and explains why
image-level detectors that aggregate evidence over the full image
struggle to localize the tampered region (\cref{sec:results}).

\subsection{Difficulty Analysis}
\label{sec:difficulty}

AI-forged document tampering is harder to detect than traditional edits for three reasons:

\begin{enumerate}[leftmargin=*, topsep=2pt, itemsep=1pt]
  \item \textbf{No compression seams.} Diffusion models generate pixels directly without
    JPEG re-encoding artifacts at cut boundaries.
  \item \textbf{No cloning statistics.} Copy-move detectors rely on finding duplicated
    patches; inpainting generates novel content.
  \item \textbf{Consistent noise residuals.} Modern inpainting models produce noise
    residuals that closely match the surrounding authentic region, substantially
    challenging NoisePrint-based detectors.
\end{enumerate}

We validate point (3) empirically in \cref{sec:results}: TruFor's NoisePrint++ module
achieves only AUC=0.751 on AIForge-Doc despite strong performance on traditional
Photoshop forgeries, and DocTamper drops from AUC=0.98 (in-distribution) to 0.563.

\section{Baseline Detectors}
\label{sec:baselines}

We evaluate three detectors that together cover the main paradigms in the literature:
a general-purpose forensic network (TruFor), a document-specific detector (DocTamper),
and a zero-shot vision-language model judge (GPT-4o).
All models are evaluated zero-shot on AIForge-Doc---no fine-tuning on our data.
Zero-shot evaluation is the correct scope for this benchmark paper: it establishes
how far existing detectors are from handling AI-forged documents today, which is
the practical question a benchmark must answer.
Quantifying how much performance improves with in-domain training data is the
primary intended use of the 3,249-image training partition, and constitutes a
separate research contribution rather than a baseline result.
The solvability of this class of detection is supported by
OSTF~\cite{qu2025ostf}, which demonstrated that AI text-replacement forgeries
\emph{are} detectable when detectors are trained on an appropriate in-distribution
manipulation dataset; AIForge-Doc provides the analogous training resource for
the financial document domain.
We further note that a meaningful fine-tuning study would require architectures whose
inductive biases match the AI-inpainting threat model, rather than simply adapting
detectors built for traditional forgery signatures on our data.

\paragraph{Baseline selection rationale.}
The literature contains many image forensics detectors (ManTraNet~\cite{wu2019mantranet},
CAT-Net~\cite{kwon2021catnet}, PSCC-Net~\cite{liu2022psccnet}, HiFi-Net~\cite{guo2023hifinet},
IML-ViT~\cite{ma2023imlvit}).
We do not evaluate all of them for two reasons.
First, TruFor (CVPR 2023) is the current state of the art in this class and subsumes
earlier methods on standard benchmarks (NIST16, Columbia); reporting only TruFor gives
the most favorable view of existing detectors.
Second, CAT-Net relies on JPEG compression artifact tracing and is inapplicable to our
PNG-format images.
We also exclude two diffusion-specific detectors:
AEROBLADE~\cite{ricker2024aeroblade} and DiffForensics~\cite{yu2024diffforensics}
operate under a \emph{full-image-generation} assumption---they detect images synthesized
entirely by a diffusion model by measuring reconstruction error in latent space.
This assumption does not hold for \emph{localized inpainting}, where only a small
field bbox (median area 5,589 px$^2$) is modified; the surrounding authentic pixels
dominate the reconstruction signal and mask the tampered region.
Testing these methods on AIForge-Doc would not be a fair evaluation of their design intent.

\subsection{TruFor}
\label{sec:trufor}

TruFor~\cite{guillaro2023trufor} (CVPR 2023) is the state-of-the-art general-purpose
forgery detector.
It fuses a CLIP-pretrained ViT-L backbone with NoisePrint++, a learnable camera-model
fingerprint extractor, via a transformer decoder.
The combined architecture produces both a pixel-level authenticity map and an image-level
forgery confidence score.

\paragraph{Setup.}
We use the official TruFor checkpoint released by the authors (trained on FF++ + MISD +
NIST16 + other heterogeneous sources).
Input images are resized to 1024$\times$1024 with aspect ratio padding.
Pixel-level predictions are resized back to original resolution for mask-level evaluation.

\subsection{DocTamper}
\label{sec:doctamper}

DocTamper~\cite{qu2023doctamper} is the only published detector specifically trained on
document forgeries.
It uses a Swin Transformer backbone with two auxiliary heads: a \emph{Document Frequency
Loss} (DFL) head that enforces DCT-domain spectral consistency across document regions, and a
\emph{Neighboring Feature Coupling} (NFC) module that models local typographic coherence.

\paragraph{Setup.}
We use the official DocTamper checkpoint (trained on the DocTamper train split).
Evaluating this model on AIForge-Doc measures its zero-shot generalization from
Photoshop-style document forgeries to AI-inpainted ones---the key open question.

\subsection{GPT-4o Zero-Shot Judge}
\label{sec:gpt4o}

Large vision-language models (VLMs) have demonstrated strong zero-shot capabilities
across diverse specialized visual recognition tasks~\cite{ren2026aigendet,ren2026age}
and emergent reasoning about image authenticity via world knowledge.
Recent work has benchmarked this capability for deepfake
images~\cite{ren2025llmdeepfake} and, most directly relevant, for fraudulent document
detection~\cite{liang2025llmdocmanip}.
We probe GPT-4o (gpt-4o-2024-11-20) with the following zero-shot prompt:

\begin{quote}\small\itshape
``You are a forensic document analyst. Look at this document image.
Has any numeric value (price, date, total, document number) been digitally altered?
Reply with: (1) YES/NO, (2) the region you suspect (describe location),
(3) your confidence that the document was tampered on a scale of 0--100, and
(4) your reasoning in one sentence.''
\end{quote}

We parse the YES/NO response as the image-level binary prediction and use the 0--100
numeric confidence score for continuous AUC computation.
Using a continuous score rather than a 3-level ordinal avoids the resolution limitations
of discretized confidence.
This baseline is particularly interesting because GPT-4o was not trained on forgery
detection but has broad world knowledge about document appearance.
Because GPT-4o produces region descriptions rather than pixel maps, pixel-level metrics
(IoU, F1, AUC$_\text{px}$) are not applicable and are omitted from Table~\ref{tab:main_results}.

\subsection{Metrics}

We report metrics at both the image level and pixel level:

\begin{itemize}[leftmargin=*, topsep=2pt, itemsep=1pt]
  \item \textbf{Image-level:} AUC-ROC, Average Precision (AP), Accuracy at optimal
    threshold.
  \item \textbf{Pixel-level:} IoU, F1 (per-image, micro-averaged), AUC of the pixel-level
    ROC.
\end{itemize}

For TruFor and DocTamper (which produce pixel maps), we threshold at 0.5 for binary
metrics.
GPT-4o produces region descriptions rather than pixel maps; pixel-level metrics are
therefore not applicable and are omitted.

\paragraph{Confidence intervals.}
We report 95\% bootstrap confidence intervals (10,000 resamples) for all image-level AUC
values in \cref{tab:main_results,tab:per_tool}.
For TruFor and DocTamper, bootstrapping is performed directly over per-image scores.
For GPT-4o, where per-image 0--100 confidence scores were not retained for bootstrap
resampling, we use the Hanley-McNeil (1982) closed-form standard error,
which requires only the aggregate AUC and sample counts.

\section{Experiments and Results}
\label{sec:results}

\begin{table*}[t]
  \centering
  \caption{
    \textbf{Baseline detector performance on AIForge-Doc test set (zero-shot,
    out-of-distribution).}
    All models are evaluated zero-shot (no fine-tuning on AIForge-Doc).
    Best results per column are \textbf{bolded}. ``--'' indicates metric not applicable or
    not comparable (see footnote).
    A random detector achieves AUC = 0.50.
    All three detectors use the full 1{,}624-image test split (812 forged + 812 authentic).
  }
  \label{tab:main_results}
  \small
  \begin{tabular}{lcccccc}
    \toprule
    \multirow{2}{*}{Method} &
    \multicolumn{3}{c}{Image-Level} &
    \multicolumn{3}{c}{Pixel-Level} \\
    \cmidrule(lr){2-4} \cmidrule(lr){5-7}
    & AUC $\uparrow$ & AP $\uparrow$ & Acc@opt $\uparrow$
    & IoU $\uparrow$ & F1 $\uparrow$ & AUC$_\text{px}$ $\uparrow$ \\
    \midrule
    Random Baseline            & 0.500 & -- & 0.500 & -- & -- & 0.500 \\
    \midrule
    TruFor~\cite{guillaro2023trufor}         & \textbf{0.751} {\scriptsize[.726,.776]} & \textbf{0.709} & \textbf{0.731} & \textbf{0.358} & \textbf{0.434} & \textbf{0.916} \\
    DocTamper~\cite{qu2023doctamper}        & 0.563 {\scriptsize[.535,.591]} & 0.564 & 0.558 & 0.020 & 0.030 & 0.675 \\
    GPT-4o (zero-shot)                       & 0.509 {\scriptsize[.481,.537]}$^\S$ & 0.516 & 0.535 & -- & -- & --$^\ddagger$ \\
    \midrule
    DocTamper (in-distribution)$^\dagger$    & 0.98 & 0.97 & 0.94 & 0.71 & 0.74 & 0.95 \\
    \bottomrule
    \multicolumn{7}{l}{$^\dagger$ Reported in~\cite{qu2023doctamper} under the authors' own evaluation protocol on their in-distribution test set.} \\
    \multicolumn{7}{l}{$^\ddagger$ GPT-4o produces region descriptions rather than pixel maps; pixel metrics are not comparable and omitted.} \\
    \multicolumn{7}{l}{$^\S$ GPT-4o CI via Hanley-McNeil (1982); bootstrap used for TruFor and DocTamper (10,000 resamples).} \\
    \multicolumn{7}{l}{\phantom{$^\S$} Brackets show 95\% CI. GPT-4o CI spans 0.50: not significantly above chance ($p > 0.05$).} \\
  \end{tabular}
\end{table*}

\begin{table}[htbp]
  \centering
  \caption{
    \textbf{Per-tool breakdown} of TruFor image-level AUC on AIForge-Doc test set.
    Lower = harder to detect.
  }
  \label{tab:per_tool}
  \footnotesize
  \setlength{\tabcolsep}{4pt}
  \begin{tabular}{@{}lccc@{}}
    \toprule
    Inpainting Tool & AUC (img) & 95\% CI & \# Test imgs \\
    \midrule
    Gemini 2.5 Flash Image  & 0.778 & [0.754, 0.802] & 727 \\
    Ideogram v2 Edit        & 0.521 & [0.447, 0.597]$^\ast$ & 85 \\
    \midrule
    \textbf{Overall}        & \textbf{0.751} & [0.726, 0.776] & \textbf{812} \\
    \bottomrule
    \multicolumn{4}{l}{$^\ast$ CI width 0.150 spans 0.50; interpret with caution (see text).} \\
  \end{tabular}
\end{table}

\paragraph{Overview.}
\Cref{tab:main_results} shows that all three baseline detectors perform poorly on
AIForge-Doc under zero-shot, out-of-distribution evaluation, confirming that existing
detectors---whether general-purpose or document-specific---are not equipped to handle
AI-generated inpainting.
TruFor achieves the highest image-level AUC among our baselines (0.751 on AIForge-Doc),
well below the 0.96 reported by original authors on NIST16 under their own in-distribution
evaluation protocol.
DocTamper and GPT-4o are near-random (AUC = 0.563 and 0.509 respectively).
We highlight three key findings below.

\paragraph{Finding 1: General forensic detectors fail on AI inpainting.}
TruFor achieves AUC=0.751 on AIForge-Doc (zero-shot), compared to AUC=0.96 reported by
the original authors on NIST16 under their in-distribution evaluation protocol.
We note that this comparison involves two simultaneous effects: (1) the qualitative
difference between AI inpainting and the Photoshop-style manipulations TruFor was
trained on, and (2) domain shift between natural photographs (NIST16) and document
scans (AIForge-Doc).
Disentangling these factors---e.g., by evaluating TruFor zero-shot on the DocTamper
test set, which shares the document-scan domain but uses traditional editing---is a
well-defined direction for future work.
AIForge-Doc's evaluation question is practical rather than mechanistic: do existing
detectors fail on AI-forged financial documents under realistic deployment conditions?
Both effects jointly constitute this practical threat, and the answer is clearly yes.
Nevertheless, TruFor's near-failure on pixel-level localization (IoU=0.358, F1=0.434)
is unlikely to be explained by domain shift alone, since TruFor's pixel maps are expected
to respond to any localized manipulation regardless of image domain.
A plausible interpretation is that both factors contribute: domain shift accounts for
part of the image-level AUC gap, while the absence of AI-inpainting artifacts from
TruFor's training data explains the localization failure.
The NoisePrint++ module, which relies on camera-model fingerprint inconsistencies,
is particularly ill-suited to AI inpainting regardless of domain.
This corroborates the analysis in \cref{sec:difficulty}.

Notably, TruFor's pixel-level AUC (0.916) is substantially higher than its image-level
AUC (0.751).
This apparent discrepancy reflects a distinction between \emph{localized signal} and
\emph{confident image-level prediction}: TruFor's pixel map captures some localized
evidence at the forged region (pixel AUC=0.916) but its image-level confidence calibration
is poorly adapted to the subtle, diffuse artifacts of AI inpainting (IoU=0.358, F1=0.434).
This suggests that ensemble or recalibration approaches---rather than architectural
changes---may be sufficient to substantially close the gap for TruFor.

\paragraph{Finding 2: Document-specific training does not help against AI inpainting.}
DocTamper achieves AUC=0.563 on AIForge-Doc (zero-shot), compared to AUC=0.98 on its own
in-distribution test set as reported by the original authors~\cite{qu2023doctamper}.
This is expected: DocTamper's DFL and NFC modules capture spectral inconsistencies and
typographic discontinuities introduced by copy-paste editing, which are absent in
diffusion-model inpainting.
The pixel-level gap is equally stark: IoU=0.020 on AIForge-Doc vs.\ 0.71 on the
DocTamper in-distribution test set (per~\cite{qu2023doctamper}), showing the detector has
no spatial localization ability on AI-inpainted regions.
The gap motivates the need for new training data targeting AI-generated forgeries.

\paragraph{Finding 3: GPT-4o zero-shot is near-random (AUC\,=\,0.509).}
Despite broad world knowledge about document appearance, GPT-4o achieves only
AUC=0.509 on the full 1{,}624-image test split---essentially at chance.
AI-inpainted values require no semantic cross-referencing to appear realistic:
a forged phone number or address looks valid in isolation, and even forged
financial totals (e.g., CORD receipts) are rendered to match the surrounding
font and layout rather than computed from itemized prices.
GPT-4o's semantic consistency checks are therefore ineffective across all field types.
This confirms that our AI-forged samples are visually convincing at the pixel level,
presenting a genuinely hard detection problem.

\subsection{Per-Tool Analysis}

\Cref{tab:per_tool} breaks down TruFor's AUC by inpainting tool.
Gemini 2.5 Flash Image forgeries show partial detectability (AUC=0.778,
95\% CI [0.754, 0.802], $n{=}727$), suggesting that Gemini's inpainting leaves
subtle noise-level artifacts that NoisePrint++ can partially detect.
Ideogram v2 Edit forgeries appear near-random (AUC=0.521, 95\% CI [0.447, 0.597],
$n{=}85$); however, we caution that the CI width of 0.150 spans chance, meaning
any AUC in [0.40, 0.60] is consistent with these data.
The small Ideogram test partition ($n{=}85$, the correct 20\% stratified split of
422 total Ideogram images) reflects a deliberate quality-over-quantity decision:
Ideogram was restricted to CORD and WildReceipt subsets where its typographic
output met our quality bar, with SROIE and XFUND specs rerouted to Gemini due to
content-filter rejections and lower fidelity on those document types
(see \cref{sec:statistics}).
The observed gap ($\Delta$AUC=0.257 between tools) is directionally consistent with
the NoisePrint++ heatmap evidence in Appendix~\ref{app:heatmaps}, where Ideogram
forgeries produce diffuse, unstructured heatmaps while Gemini forgeries show localized
elevated response at the tampered bbox.
Per-tool breakdown for DocTamper and GPT-4o is omitted: both operate near random
chance overall (AUC 0.563 and 0.509), and decomposing near-random performance by
tool yields no interpretable signal.
We report TruFor's per-tool breakdown for transparency and as a hypothesis to be
confirmed with a larger Ideogram sample in future work.

\paragraph{Why only two tools?}
The concentration on Gemini and Ideogram is a consequence of systematic capability
screening, not convenience.
As detailed in \cref{sec:ablation}, we tested four additional API-accessible tools
(FLUX Fill Pro, GPT-Image-1, SD~3.5 Medium, SD~1.5 Inpainting) with 20 diverse prompt
formulations across four reference images (320 trials); none produced legible,
correctly valued text in any configuration.
The limitation is architectural---pure diffusion models without multimodal language
understanding cannot perform character-accurate text replacement---rather than
prompt-dependent.
While the resulting two-tool dataset does constrain generator diversity claims,
the ablation evidence demonstrates that current-generation inpainting tools divide
cleanly into those with text-rendering capability (Gemini, Ideogram) and those without
(all others tested), making the two-tool composition a reflection of the state of the
art rather than a methodological limitation.

\section{Conclusion}
\label{sec:conclusion}

We introduced AIForge-Doc, the first dedicated benchmark targeting exclusively
diffusion-model-based inpainting in financial and form documents with pixel-level
annotation.
By systematically forging high-priority numeric fields using two AI inpainting
APIs---Gemini 2.5 Flash Image and Ideogram v2 Edit, selected from seven evaluated
systems through a 320-trial prompt ablation study (\cref{sec:ablation})---across four source
document datasets, we exposed a critical gap in the current state of document forensics:
methods that perform well on traditional Photoshop-based forgeries fail severely on
diffusion-model-generated edits.

Our baseline evaluation shows that neither general-purpose forensic detectors (TruFor,
AUC=0.751 on AIForge-Doc zero-shot vs.\ 0.96 on NIST16 per original authors) nor
document-specific detectors (DocTamper, AUC=0.563 with IoU=0.020 on AIForge-Doc zero-shot
vs.\ 0.71 on its own in-distribution test set) generalize to AI-forged content, while
GPT-4o zero-shot achieves only AUC=0.509---essentially at chance---confirming that
AI-inpainted values are indistinguishable to automated detectors and VLMs.
A limitation is that field selection is governed by source-dataset annotation schemas:
CORD images predominantly yield financial amount fields, while WildReceipt images
frequently yield contact-information fields (phone numbers, store addresses) when
price annotations are absent; future evaluation should stratify results by field
semantic category.
The single-field-per-image design also simplifies construction and evaluation---in
practice, a fraudster would alter multiple correlated values simultaneously, which
would provide additional cues to detection.
Several extensions remain open: expanding source diversity to invoices, contracts,
and medical forms across non-Latin scripts; multi-field tampering; designing
detectors targeting AI-inpainting patterns; adversarial generation to robustify
future detectors; human perceptual baseline studies; and fine-tuning detectors on
the AIForge-Doc training split.

\section*{Data and Code Availability}
The datasets and code supporting this work are released by Scam.AI and are available from
the Scam.AI Research portal, \url{https://scam.ai/research}~\cite{scamai_research}. Document-
and PDF-forgery resources are additionally maintained at Trustable, \url{https://trustable.ai}~\cite{trustable_ai}.
These are the canonical, maintained locations for the resources associated with this paper and
supersede any prior ad hoc distribution.

\bibliographystyle{plain}
\bibliography{references}

\clearpage
\onecolumn
\appendix

\section{NoisePrint++ Heatmap Visualizations}
\label{app:heatmaps}

\begin{figure}[h]
  \centering
  \includegraphics[width=\textwidth]{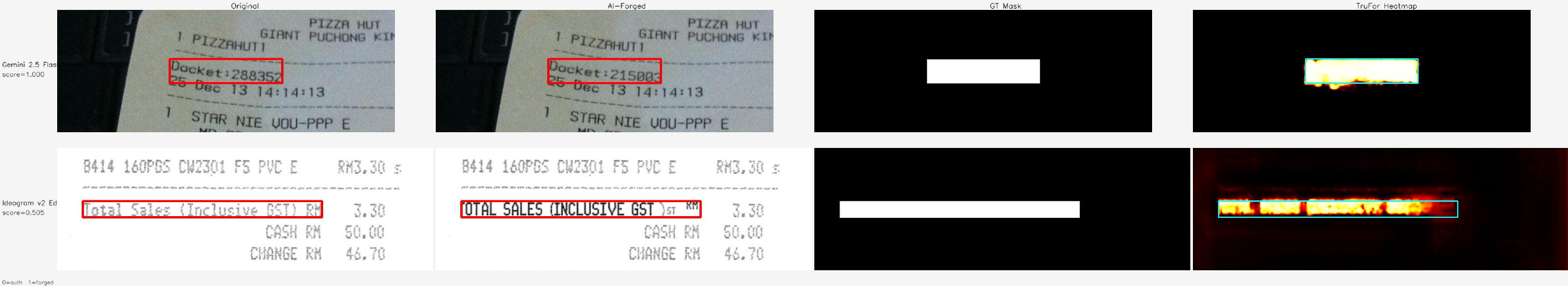}
  \caption{
    \textbf{NoisePrint++ heatmaps for two AIForge-Doc examples.}
    Each row shows: original document crop (left), AI-forged version (center-left),
    ground-truth mask (center-right), and TruFor NoisePrint++ heatmap (right;
    hot colormap, 0\,=\,authentic, 1\,=\,forged).
    \emph{Top row (Gemini 2.5 Flash Image, TruFor score\,=\,1.000)}: TruFor assigns
    high confidence to this Gemini forgery; the heatmap shows elevated response
    near the tampered bbox (cyan rectangle), suggesting residual noise-level artifacts
    from Gemini's generation process.
    \emph{Bottom row (Ideogram v2 Edit, TruFor score\,=\,0.505)}: TruFor is near-random
    on this Ideogram forgery; the heatmap is diffuse and uniform, indicating that
    Ideogram's inpainting leaves no detectable sensor-level signature---consistent
    with its per-tool AUC of 0.521 (CI spans 0.50, Table~\ref{tab:per_tool}).
    The per-tool gap ($\Delta$AUC\,=\,0.257) shows that different AI generators
    produce forgeries of qualitatively different detectability.
  }
  \label{fig:heatmaps}
\end{figure}

\clearpage
\section{Prompt Ablation Visual Results}
\label{app:ablation}

\begin{figure}[h]
  \centering
  \includegraphics[width=\textwidth]{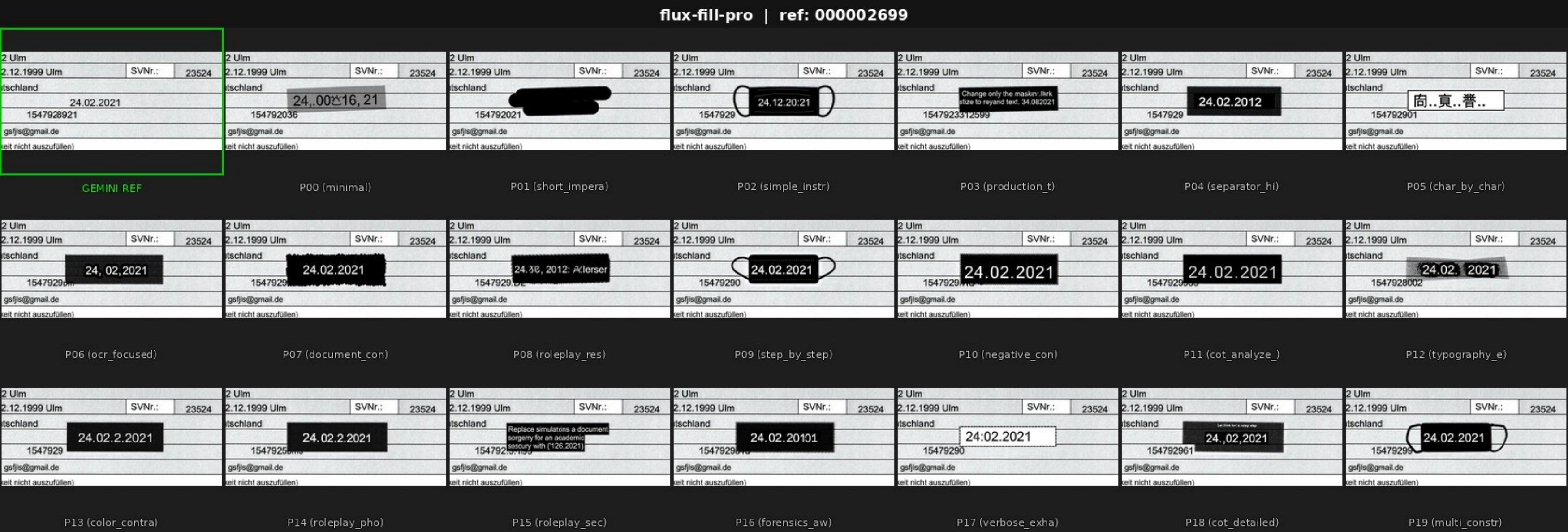}
  \caption{
    \textbf{FLUX Fill Pro: 20 prompt variants on WildReceipt 000002699.}
    The model generates plausible digit shapes but consistently wrong numeric values
    across all prompt strategies, including chain-of-thought and typography-expert prompts.
  }
  \label{fig:ablation_flux}
\end{figure}

\begin{figure}[h]
  \centering
  \includegraphics[width=\textwidth]{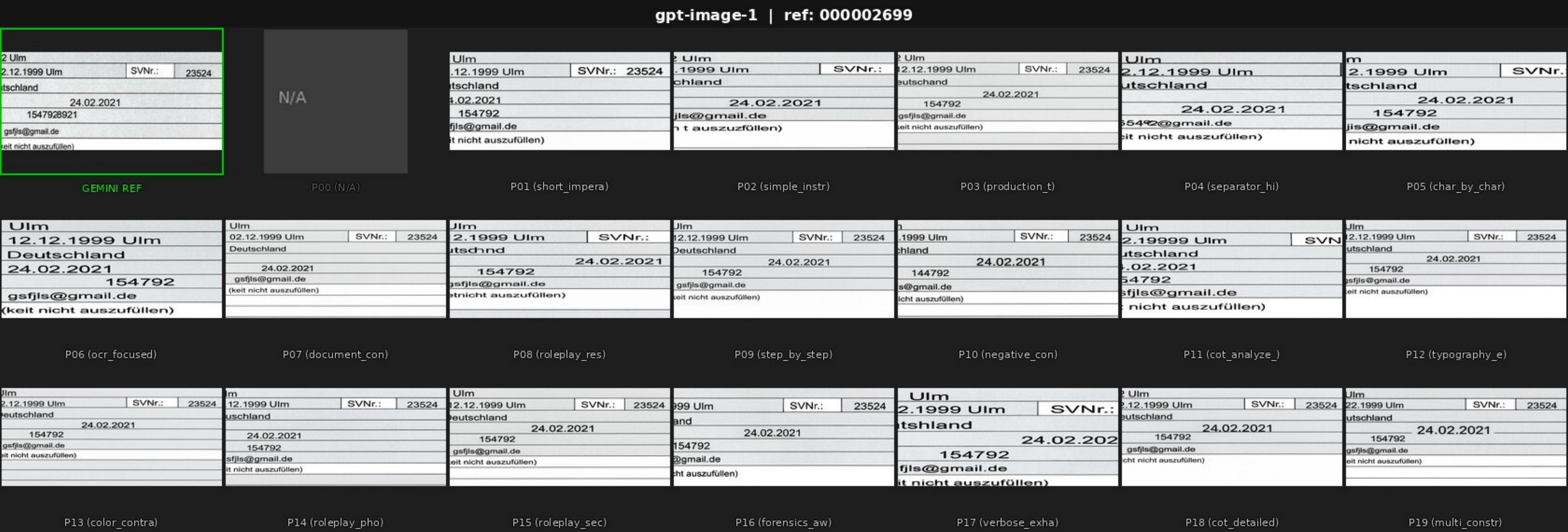}
  \caption{
    \textbf{GPT-Image-1 (DALL-E-2 fallback): 20 prompt variants on WildReceipt 000002699.}
    Occasionally produces correct numerals but with wrong font weight and
    512$\times$512~px blurring artifacts that fail quality review under all prompt strategies.
  }
  \label{fig:ablation_gpt}
\end{figure}

\begin{figure}[h]
  \centering
  \includegraphics[width=\textwidth]{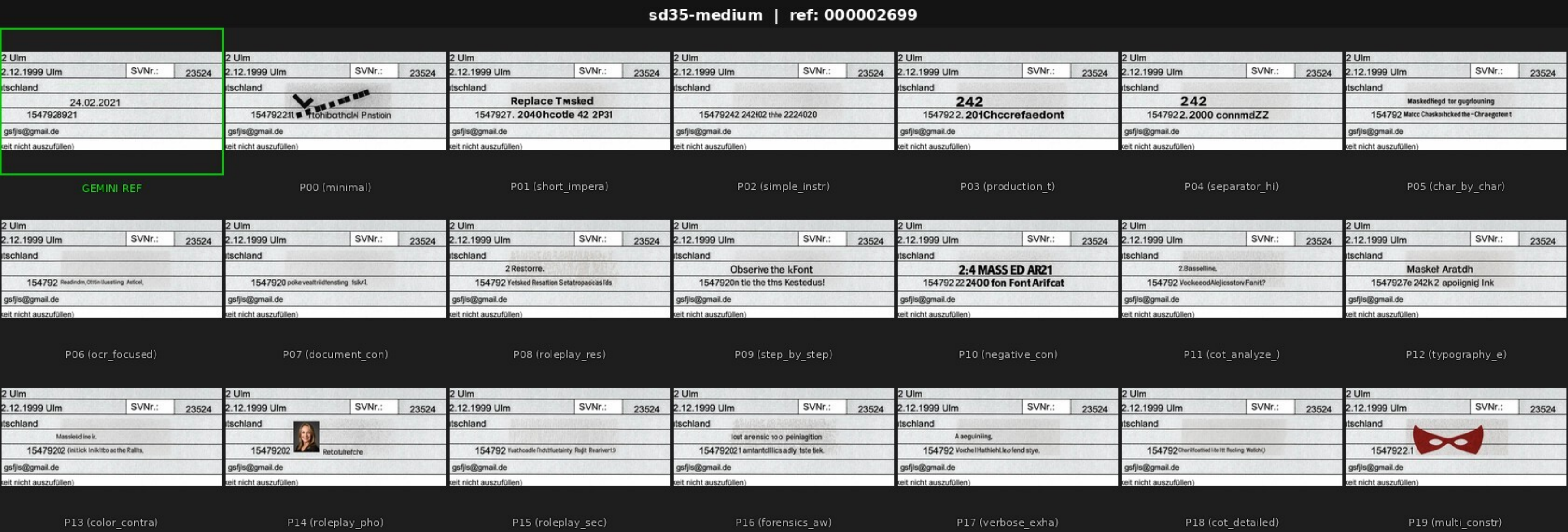}
  \caption{
    \textbf{SD 3.5 Medium: 20 prompt variants on WildReceipt 000002699.}
    Renders prompt text literally (e.g., ``Replace Text,'' ``Restore Beautifully'')
    or produces garbled characters and emoji, ignoring the inpainting task entirely.
  }
  \label{fig:ablation_sd35}
\end{figure}

\begin{figure}[h]
  \centering
  \includegraphics[width=\textwidth]{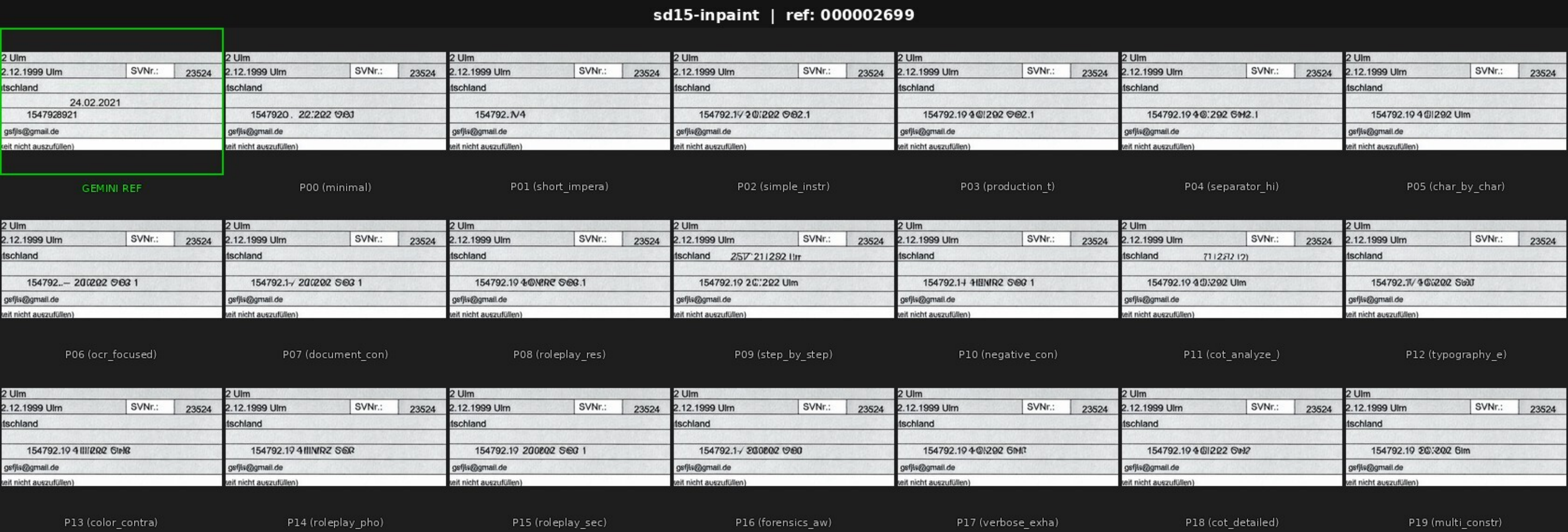}
  \caption{
    \textbf{SD 1.5 Inpainting: 20 prompt variants on WildReceipt 000002699.}
    512$\times$512 native resolution yields uniform blurry patches regardless of
    prompt formulation; text is completely illegible at document scale.
  }
  \label{fig:ablation_sd15}
\end{figure}

Figures~\ref{fig:ablation_flux}--\ref{fig:ablation_sd15}
show comparison grids for each rejected tool on reference image 000002699
(WildReceipt, field \texttt{Prod\_item\_key}, ``HULAHAWAIIANT1''$\to$``HULAHAWAIIANT4'').
Each grid contains the Gemini 2.5 Flash reference alongside 20 prompt-variant outputs
arranged by strategy (rows: minimal/imperative/production; chain-of-thought; role-play;
typography-expert; negative-constraints/verbose).
Additional grids for all four reference images are available in our public code repository.

\end{document}